\documentclass[runningheads]{llncs}
\usepackage[T1]{fontenc}
\usepackage[misc]{ifsym}
\usepackage{amsmath,amssymb}
\usepackage{float}
\usepackage{placeins}

\usepackage{graphicx}
\begin{document}
\title{SCPainter: A Unified Framework for Realistic 3D Asset Insertion and Novel View Synthesis}
\titlerunning{SCPainter: Realistic 3D Asset Insertion and NVS}
%
\author{Paul Dobre\inst{1} \and
Jackson Cooper\inst{1} \and
Xin Wang\inst{1}
\and Hongzhou Yang\inst{1}\textsuperscript{(\Letter)}}
\authorrunning{P. Dobre et al.}
%
\institute{Department of Geomatics Engineering, University of Calgary, Calgary, Alberta, Canada \\
\email{honyang@ucalgary.ca}}
\maketitle
\begin{abstract}
3D Asset insertion and novel view synthesis (NVS) are key components for autonomous driving simulation, enhancing the diversity of training data. With better training data that is diverse and covers a wide range of situations, including long-tailed driving scenarios, autonomous driving models can become more robust and safer. This motivates a unified simulation framework that can jointly handle realistic integration of inserted 3D assets and NVS. Recent 3D asset reconstruction methods enable reconstruction of dynamic actors from video, supporting their re-insertion into simulated driving scenes. While the overall structure and appearance can be accurate, it still struggles to capture the realism of 3D assets through lighting or shadows, particularly when inserted into scenes. In parallel, recent advances in NVS methods have demonstrated promising results in synthesizing viewpoints beyond the originally recorded trajectories.  However, existing approaches largely treat asset insertion and NVS capabilities in isolation. To allow for interaction with the rest of the scene and to enable more diverse creation of new scenarios for training, realistic 3D asset insertion should be combined with NVS. To address this, we present SCPainter (Street Car Painter), a unified framework which integrates 3D Gaussian Splat (GS) car asset representations and 3D scene point clouds with diffusion-based generation to jointly enable realistic 3D asset insertion and NVS. The  3D GS assets and 3D scene point clouds are projected together into novel views, and these projections are used to condition a diffusion model to generate high quality images. Evaluation on the Waymo Open Dataset demonstrate the capability of our framework to enable 3D asset insertion and NVS, facilitating the creation of diverse and realistic driving data.

\end{abstract}
\section{Introduction}
Autonomous driving has progressed rapidly, driving demand for large amounts of high-quality data that span diverse driving scenarios. Particularly, end-to-end autonomous driving models have emerged as a promising training-based approach, learning driving behavior directly from data [3]. Existing approaches to training autonomous driving models can be categorized into two paradigms: open-loop learning from recorded data and closed-loop learning through simulation or interactive environments. Open-loop data provides the benefit of real sensors captured from expert trajectories but does not allow the agent to explore in the scene, limiting training scenarios. Closed-loop data or simulation allows for diverse scenarios for training, but results in poor generalization to real data due to training on simulated sensors. As a result, there is an increasing demand for simulation that maintains the realism of open-loop recordings, while providing the diverse and controllable aspects of closed-loop simulation.

\indent To bridge the gap between realism and controllability, recent approaches have explored novel view synthesis (NVS) and 3D reconstruction techniques to generate interactive driving environments from real-world video recordings [5, 27]. However, these methods can significantly degrade in quality when novel viewpoints deviate from the original captured poses.  Methods like ReconDreamer, GEN3C and FreeVS [14, 16, 21], seek to improve the quality of extrapolated views by introducing pre-trained generative models [1, 20] conditioned on geometric control signals such as projected lidar points, generated depth, or deteriorated Gaussians. Another avenue explored in the pursuit of realistic simulation, is improved 3D asset reconstruction from driving videos. Liu et al. [13] explores the using Fibonacci sphere sampling to obtain initial Gaussian points, and a tri-plane representation to decode the Gaussian attributes, while [11] explores fine-tuning a large 3D asset reconstruction model on driving scene specific data and [30] uses a 3D latent diffusion model trained on driving scenes to generate 3D driving assets. However, due to limitations in current 3D asset reconstruction methods and sparsely captured views of the objects, the reconstruction quality can significantly degrade. Furthermore, even with high quality reconstruction, when 3D assets are placed back into the scene, they lack realistic lighting and shadows. This negatively impacts the training of autonomous driving models as inserted 3D assets have unrealistic appearance compared to real-world captured cars. Recently, R3D2 [11] explores how to improve the realism of asset insertion by training a diffusion model to recover realistic lighting, shadows and improved detailed structure for inserted 3D car assets. This has significant benefit as it allows for generation of diverse driving scenarios with realistic cars that were not present in the original recordings, improving the diversity of training data. However, while these approaches improve asset realism, they do not support simulating novel camera trajectories or dynamic interactions with the newly introduced assets. At the same time, NVS approaches are not designed to realistically integrate newly inserted 3D assets. As a result, existing methods do not provide a unified solution that simultaneously enables realistic 3D asset rendering and controllable NVS.

\indent We propose SCPainter (Street Car Painter), a framework for jointly performing 3D car asset insertion and NVS in driving scenes. The framework enables high quality integration of newly inserted assets under controllable novel camera trajectories, addressing the lack of unified methods for augmenting driving data. Our approach represents scene geometry using colorized point clouds obtained by unprojecting estimated depth, and represents inserted asset geometry using 3D GS. These two representations can be projected to novel camera trajectories, providing geometric conditioning for video generation. A video diffusion model is then fine-tuned to transform these renderings into visually consistent novel-view videos, correcting depth artifacts, filling missing regions, and integrating inserted assets while maintaining temporal consistency.

\indent We evaluate SCPainter across multiple driving video generation tasks, including standard asset insertion, novel view synthesis, and their simultaneous combination. Results demonstrate accurate rendering of novel viewpoints, realistic integration of inserted assets, and effective coupling of both objectives to generate diverse driving scenarios. Experiments on the Waymo Open Dataset validate the effectiveness of our approach.

\section{Related Work}
\textbf{3D Asset Reconstruction and Insertion.} 3D reconstruction of assets from driving video has received significant attention. Prior approaches leverage diffusion models to generate complete 3D assets from sparse and occluded imagery [9, 30], focusing specifically on 3D car asset reconstruction from street scenes. Generalized models leverage structured 3D latents to achieve strong reconstruction performance that achieves comparable performance to specialized methods for street asset reconstruction. However, these methods still struggle with fine geometric detail and when naively inserted in scenes, 3D assets lack scene-specific illumination and shadow interactions, resulting in visually inconsistent simulations. Several approaches address the realistic insertion of assets like cars with image diffusion models, demonstrating strong performance [10, 11, 17], however these focus exclusively on the asset insertion case. Most similar to our approach is R3D2 [11], which uses image-to-image diffusion to realistically render naively inserted cars into street scenes. However, this approach processes frames independently, which can result in temporal inconsistency. Furthermore, it does not address NVS combined with the 3D asset insertion, limiting the potential of novel driving simulations that can be generated. In this work, we present a framework for coupling the insertion of 3D assets into scenes with NVS, enabling diverse and realistic driving scenario generation.

\textbf{Novel View Synthesis.} Recently, significant progress has been achieved in NVS through advancements in 3D reconstruction and diffusion models. 3D Gaussian splatting [8] represents 3D scenes using Gaussian primitives optimized for 3D position, opacity ($\alpha$), anisotropic covariance, and spherical harmonic coefficients. Tile-based rendering allows for efficient training and real-time rendering of the 3D scene. 3D GS, however, was initially designed for static and bounded scenes. Some researchers have extended 3D GS to driving scenes, modelling large street scenes with dynamic objects present [5, 27, 32]. However, these approaches begin to significantly degrade in quality as the trajectory deviates from the recorded driving trajectory. To address this, several researchers have proposed to leverage diffusion models to restore degraded Gaussians in novel trajectories [6, 14]. The 3D GS from the original trajectory is rendered into the novel view. This degraded rendering is then passed into a diffusion model as conditioning, and a restored novel trajectory is generated. The 3D GS can then be re-optimized with these generated views, and this process can be done progressively until the desired novel view trajectory is reached. Other approaches leverage colourized points (LiDAR points or unprojected depth estimates) as a geometric prior that can be projected into novel views and passed as conditioning into the diffusion model [16, 21]. The diffusion model is trained to fill in areas missing projected points and to resolve artifacts resulting from the projection of imperfect depth estimation/measurements

\section{Methodology}
As seen in Fig. 1, we present SCPainter, a generative framework that is trained to realistically integrate inserted 3D assets into existing scenes while simultaneously enabling precise, consistent and high-quality NVS. In section 3.1, we provide background for Stable Video Diffusion (SVD) [1]. In section 3.2, we outline the methods used to reconstruct both the 3D asset and 3D scene. In section 3.3, we describe the rendering of the 3D GS asset and colorized point cloud, and how it is used as conditioning by the SVD model. Then in section 3.4, we describe the training process for the model.

\begin{figure}
\includegraphics[width=\textwidth]{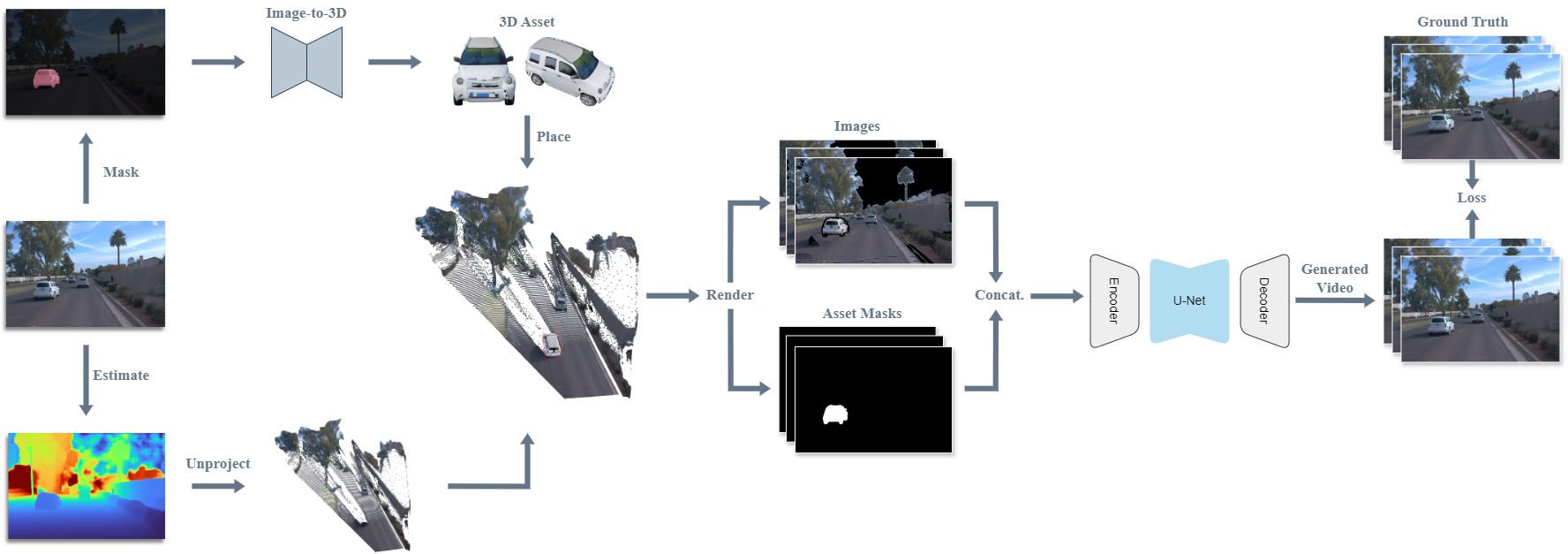}
\caption{Method Framework for SCPainter. Assets are masked out from the driving video and reconstructed with Amodal3R [24]. The reconstructed 3D GS asset is inserted into the unprojected predicted depth from VGGT [22]. This is then rendered as video into the user requested trajectory which is fed into the SVD model [1] to generate a photorealistic video of the user trajectory.} \label{fig1}
\end{figure}

\subsection{Background: Stable Video Diffusion}
Our generative model is based on Stable Video Diffusion (SVD) [1], which encodes images from a video into a latent space, with the SVD being trained and performing inference in this latent space. This enhances the efficiency of the generation and focuses modeling capacity on meaningful semantic and dynamic structure in the image. Given an RGB video $x \in \mathbb{R}^{T \times 3 \times H \times W}$, where T is the number of frames of size $H \times W$, a pre-trained VAE encoder $\mathcal{E}$ encodes the video frames into latent space $\mathbf{z} = E(x) \in \mathbb{R}^{T \times C \times h \times w}$, where $h = \frac{H}{8}$ and  $w = \frac{W}{8}$. The U-Net of the diffusion model is then trained in this latent space to model a data distribution via learning to predict noise that is iteratively added. To train the model, noisy versions of data $x_0$ are constructed by adding noise $\epsilon$ sampled from the Gaussian distribution $N(0,I)$, with the noise schedule parametrized by ${\alpha}_\tau$ and ${\sigma}_\tau$ yielding $x_\tau = \alpha_\tau x_0 + \sigma_\tau \varepsilon$. The diffusion time $\tau$ is sampled from the distribution $p_{\tau}$. The parameters $\theta$ of the diffusion model $f(\theta)$ are optimized to predict the noise that was added according to the noise schedule:

\begin{equation}
\mathbb{E}_{x_0 \sim p_{\text{data}}(x),\, \tau \sim p_\tau,\, \varepsilon \sim \mathcal{N}(0, I)}
\left[ \left\| f_\theta(x_\tau; c, \tau) - y \right\|_2^2 \right]
\end{equation}

\noindent where $c$ is the optional condition and $y$ is the target. After the model is trained, it can be iteratively applied to a sample of Gaussian noise to produce a sample of data that is then decoded with a pre-trained VAE decoder $\mathcal{D}$ resulting in the final video $\mathbf{x}_{final} =\mathcal{D}(\mathbf{z})$.

\subsection{3D Asset and Scene Reconstruction}
\textbf{3D Asset Reconstruction.} To achieve asset insertion, we require complete and geometrically accurate 3D assets. For this purpose, we use Amodal3R [24], a conditional 3D generative model designed to reconstruct occluded 3D objects from partial image observations. As input, Amodal3R takes in an object-focused crop of the image, and masks corresponding to the observed portion of the object and the occluding object(s). The output is a full 3D GS model of the asset. To obtain the necessary object specific masks, we employ foundational vision-language and video segmentation models [12, 15]. Overlapping masks for the object of interest are taken as the foreground occlusion masks, indicating unobserved parts of the object that were occluded in the image. The generated 3D asset composed of Gaussian primitives, can then be inserted into the scene and rendered to image. To ensure realistic appearance and seamless integration, we require high quality 3D GS assets to insert into the scene. Visually inconsistent or noisy insertions can bias generative models and reduce realism in synthesized views. While Amodal3R [24] achieves strong performance, due to the nature of reconstructing occluded objects from 2D images, Amodal3R can hallucinate incorrect object shape and appearance. Similar to R3D2 [11], we perform a two-stage filtering process for the generated 3D assets. We first use the asset-specific bounding box to filter assets that significantly deviate from the original object dimensions. The remaining 3D assets are manually inspected, and those that appear unrealistic or deviate significantly from the original object are filtered.	

\textbf{3D Scene Reconstruction.} To enable NVS, similar to Gen3C [16], we rely on a 3D point representation for the scene and objects that were not replaced with 3D GS versions. Depth estimation and scene reconstruction has achieved significant progress across diverse domains [2, 22, 28, 29]. Therefore, leveraging the accuracy and speed of VGGT [22], we unproject colourized points from the image depth estimation, and use this as scene-level geometric conditioning. The colourized points can then be projected into the desired camera trajectory, yielding strong geometric and colour priors. VGGT is used to simultaneously recover the depths and poses for all the frames in a segment. The reconstructed 3D assets are integrated into the VGGT-derived point clouds at each frame, allowing both scene geometry and asset geometry to be jointly projected into the novel view. The joint projection is then used to condition the generation of the SVD model.

\subsection{Rendering and Conditioning with Geometry}
The unprojected colourized points along with the 3D GS assets placed into the scene can be efficiently rendered together into new camera trajectories. The colourized point rendering results in two outputs, an RGB image $I$ containing the projected colourized points, and a mask $M$, denoting the pixels that did not receive any projected points. The 3D GS asset is projected together with the colourized points into image $I$, and mask $M_a$ is extracted denoting the rendering of the 3D GS into the image. The SVD model must inpaint the regions of the image identified by the mask and repair any artifacts that resulted from the projection of the point clouds. Furthermore, the model must render the 3D GS asset inserted in the scene with realistic shadows and lighting. For a given set of camera $C=(C^1,...,C^T)$, the points unprojected from each observed viewpoint along with the inserted 3D GS asset are rendered into the new camera poses providing rendered images $(I^{1,v},...,I^{T,v})$ and masks $(M^{1,v},...,M^{T,v})$. In addition, we extract masks denoting the rendering of the 3D GS into each observed viewpoint as $\left( M_a^{1,v}, \ldots, M_a^{T,v} \right)$. This provides additional asset-aware spatial conditioning, allowing the model to distinguish regions corresponding to the inserted 3D GS asset from background projected points, and to accurately model shadowing and illumination effects induced by the asset.

\indent This yields the images $I^v \in \mathbb{R}^{T \times 3 \times H \times W}$, projected point masks $M^v \in \mathbb{R}^{T \times 1 \times H \times W}$, and projected 3D GS asset masks $M_a^v \in \mathbb{R}^{T \times 1 \times H \times W}$. These outputs are then used as the conditioning for the video diffusion model for the generation of the novel driving scenario. For the forward computation of the Stable Video Diffusion model, the rendered video $I^v$ is first concatenated with the asset masks $M_a^v$. Then they are encoded with the pre-trained VAE encoder, acquiring the latent video $z^v=\mathcal{E}(I^v)$. Like GEN3C [16], the masks $M^v$ are down sampled to the same dimensions as the latents $z^v$ to size $\frac{H}{8} \times \frac{W}{8}$, aligning the masks with the latent dimension. Element-wise multiplication is then performed between the down-sampled masks $M^{v'}$ and the latent $z^v$, yielding masked latent $z^{v'}$. This effectively zeros out the latent in regions that no points were projected and rendered. Prior to the down sampling and element-wise multiplication, the asset masks $M_a^v$ are also composited into the projected point masks to ensure the area of the projected 3D GS asset is not also zeroed out. For training, this masked latent is concatenated with the noisy version of the target video $x$ in latent space $z_\tau = \alpha_\tau x + \sigma_\tau \varepsilon$ along the channel dimension and input into the SVD model.

\subsection{Model Training}
The SVD model is trained with two objectives: to realistically integrate inserted 3D assets into the existing scene, and to refine the colourized points projected into the novel view trajectory. 3D car assets suitable for reconstruction are selected from the WOD [18] and reconstructed with Amodal3R [24]. For the training of realistic 3D asset insertion, we then remove selected assets that were reconstructed from the scene and replace them with the generated 3D GS asset. The 3D asset is then aligned with the orientation and position of the originally observed asset in the target frame, using the WOD object-level 3D bounding boxes. This effectively creates training pairs for enhancing the realism of inserted 3D assets with scene specific lighting, shading, and shadows between the generated and real asset. For the NVS aspect of training, we project points from past and future frames into the current target frame as a data augmentation to simulate the viewpoint transformation. This tasks the video diffusion model to recover the current target frame from the projected points from nearby frames. The result is a rendered image with regions of sparse projections, regions empty of points, regions with artifacts from imperfect depth estimation, and projections of the generated 3D GS assets. Thus, the model is trained to recover clean images from the rendered point and 3D GS asset projections, effectively formulating the problem as an image rectification task.

\indent The training target of SCPainter is to recover the sampled viewpoints based on the constructed renderings containing the inserted 3D assets and points projected from neighbouring frames. For training, the corresponding sequence of ground truth target views are sampled along the viewpoint sequence. The ground truth camera views are encoded as a target video latent representation with the frozen VAE encoder as $\mathcal{E}(x)$, yielding target $y \in \mathbb{R}^{T \times C \times h \times w}$. The first frame in the sequence is also encoded with the CLIP [19] model as an additional condition. For training, the CLIP condition and rendered image of the scene points and 3D GS assets are randomly dropped with a probability of $15\%$. For inference we use our trained video diffusion model to iteratively denoise latent code z instantiated as Gaussian noise, conditioned on the rendered points and 3D GS asset. The denoised latent code is then decoded using the pre-trained VAE decoder $\mathcal{D}$, yielding the final video.

\section{Experiments}
In this section, we conduct experiments on the WOD to evaluate the performance of our model in realistically inserting the reconstructed 3D GS assets and in NVS.

\textbf{Dataset.} The experiments are performed on the WOD dataset, a large-scale multimodal camera-LiDAR autonomous driving dataset [18]. It is comprised of 1000 scenes, with each scene having 200 frames. We evaluate our method with 10 selected driving sequences. The driving sequences are selected to allow space for asset insertion and for the simulation of novel trajectories by moving the camera laterally.

\textbf{Model.} The proposed SCPainter framework is implemented based on Stable Video Diffusion (SVD) [1]. The model is initialized from the SVD checkpoint. The model is trained on the WOD training set, excluding the segments selected for validation. The NVS is simulated for training by randomly sampling which frames from the adjacent ±8 frames will have their points projected into the target view. The model is trained using 4 H100 GPUs for 30,000 iterations with a batch size of 8. 

\textbf{Evaluation.} We first test our model on the standard re-insertion of the reconstructed assets at their original positions, comparing to naïve insertion (Sec. 4.1). We also assess the capability of our model in NVS, evaluating its ability to generate photorealistic and geometrically consistent novel views (Sec. 4.2). Lastly, we assess our model in the simultaneous asset insertion and NVS case (Sec 4.3). Without explicit ground truth, we evaluate the realism of the generated video with Fréchet Inception Distance (FID) [7]. The FID metric compares the overall image distribution between synthesized images on novel trajectories and ground truth images on recorded trajectories.

\subsection{Standard Insertion of Assets}
In Table 1, we evaluate the performance of SCPainter on the task of realistically inserting the 3D assets into scenes. This result is also shown qualitatively in Figure 2, demonstrating the capability of SCPainter to insert the 3D assets into scenes. Assets are selected from a validation set and inserted into scenes different from those in which they were originally generated. This setup evaluates the model’s ability to realistically insert assets into novel, previously unseen scene contexts. From the quantitative and qualitive results, our framework can understand global scene lighting and realistically cast shadows for the new assets. The insertion is compared to the naïve insertion where the asset is simply placed in the scene and not passed through the diffusion model. As can be seen, SCPainter reduces the gap in realism between the inserted car asset and other cars in the scene. It effectively integrates the new asset into the scene while preserving global scene harmony in terms of lighting, shading, and visual coherence.

\begin{table}
\centering
\caption{Results from asset insertion with SCPainter. Metrics are reported on car-centric crops denoted as FID-C.}
\label{tab1}

\setlength{\tabcolsep}{24pt} 
\renewcommand{\arraystretch}{1.2} 

\begin{tabular}{|c|c|}
\hline
Insertion Type & FID-C$\downarrow$ \\
\hline
Naive Insertion & $35.87$ \\
Ours & $16.14$ \\
\hline
\end{tabular}
\end{table}

\begin{figure}[H]
  \centering
  \includegraphics[width=\textwidth]{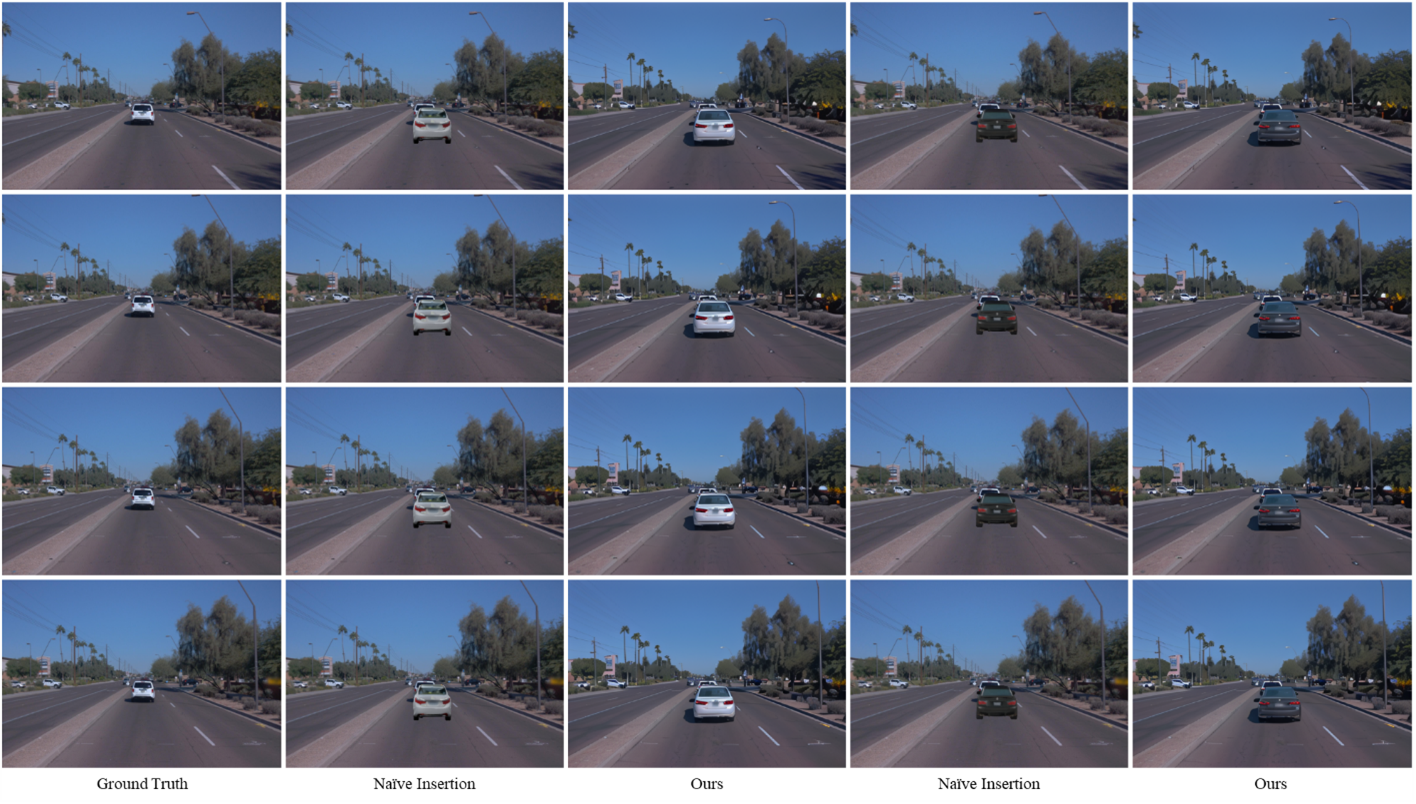}
  \caption{Qualitative results from SCPainter on asset insertion. SCPainter effectively integrates the asset into the scene with realistic shadows and lighting.}
  \label{fig1}
\end{figure}

\subsection{Novel View Synthesis}
We report FID metrics for synthesized novel views without asset insertion and compare against prior NVS methods. This evaluation demonstrates that our method preserves strong NVS performance while additionally enabling realistic asset insertion. We assess our method against previous NVS methods by generating novel trajectories with lateral camera shifts. Shifts of 2m and 3m are performed laterally relative to the heading direction of the ego vehicle. FID metrics on the novel trajectories for the different camera shifts are reported in Table 2. The results from the different methods are also visualized in Figure 3 providing a qualitative comparison between the methods. Across the shift magnitudes, our method consistently matches or outperforms the previous NVS methods in terms of FID. The 3D scene point clouds better maintain their structure and appearance under large view changes in comparison to 3D GS. Furthermore, the unprojected 3D scene point clouds from VGGT [22] are denser than LiDAR, reducing the amount of content the diffusion model must hallucinate. Together, these properties enable accurate and reliable geometric conditioning, leading to strong NVS performance. This demonstrates that the added asset insertion capability does not come at the cost of degraded NVS quality. Furthermore, our approach is LiDAR-free and uses only camera images, without sacrificing NVS quality. 

\begin{table}[!t]
\centering
\caption{Comparison with NVS methods on novel trajectories. The y-axis is defined laterally relative to the forward direction of the ego vehicle.}
\label{tab2}

\setlength{\tabcolsep}{24pt}
\renewcommand{\arraystretch}{1.2}

\begin{tabular}{|c|c|c|}
\hline
\rule{0pt}{5ex}NVS Methods &
\shortstack{$y \pm 2.0\,\mathrm{m}$\\FID$\downarrow$} &
\shortstack{$y \pm 3.0\,\mathrm{m}$\\FID$\downarrow$} \\
\hline
OmniRe [4] & $51.18$ & $64.74$ \\
FreeVS [21] & $18.84$ & $22.19$ \\
Ours & $18.43$ & $21.93$ \\
\hline
\end{tabular}
\end{table}

\begin{figure}[!t]
\centering
\includegraphics[width=\textwidth]{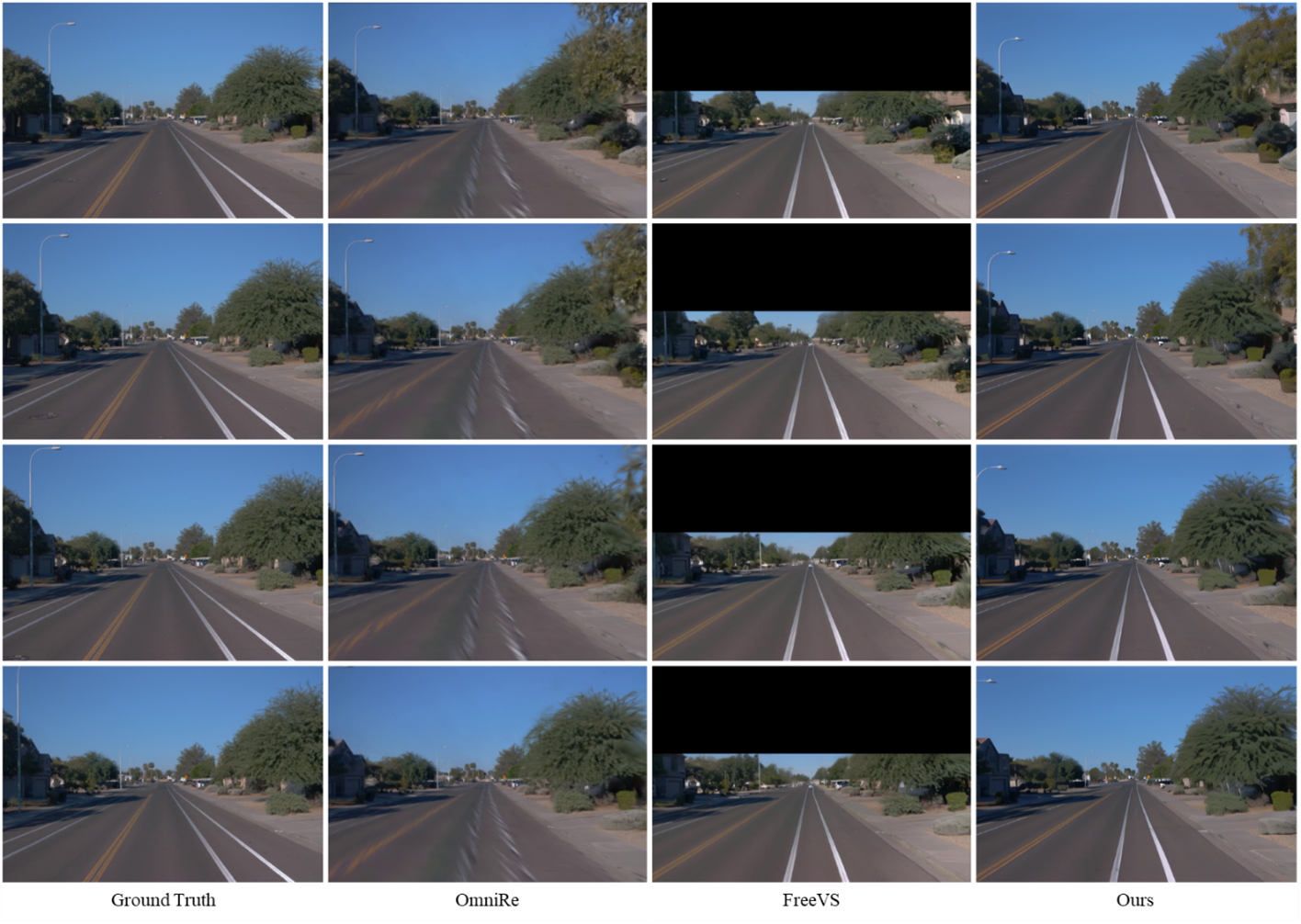}
\caption{Visual comparison between the NVS methods and ours. Viewpoint is shifted 2.0m to the right of the original camera trajectory. FreeVS does not generate sky regions in the images.} \label{fig3}
\end{figure}

\FloatBarrier
\subsection{Unified 3D Asset Insertion and Novel View Synthesis}
We evaluate SCPainter on the task of simultaneously inserting novel 3D assets while performing NVS along unseen camera trajectories. In this setting, assets are inserted into the scene and the camera is moved laterally from the original driving trajectory, requiring the model to jointly reason about asset placement and the novel view scene coherence over time. Quantitative results for this combined task are reported in Table 3, while qualitative comparisons are shown in Figure 4. The results demonstrate that SCPainter can maintain photorealistic asset integration under novel viewpoints, preserving consistent lighting, shadows, and geometric alignment as the camera moves. SCPainter generates temporally stable frames of both the asset and scene. The inserted assets remain spatially anchored to the scene and exhibit coherent shading and shadow behavior across time. These results demonstrate strong temporal consistency and visual continuity, showing that SCPainter effectively couples asset insertion with NVS while preserving realism and scene coherence.

\begin{table}
\centering
\caption{Results from simultaneous asset insertion and NVS reported with FID on the whole image. The viewpoint is shifted 2.0m in combination with asset insertion.}
\label{tab3}

\setlength{\tabcolsep}{24pt} 
\renewcommand{\arraystretch}{1.2} 

\begin{tabular}{|c|c|}
\hline
Insertion Type & FID$\downarrow$ \\
\hline
Naive Insertion & $32.03$ \\
Ours & $22.43$ \\
\hline
\end{tabular}
\end{table}

\begin{figure}[H]
\centering
\includegraphics[width=\textwidth]{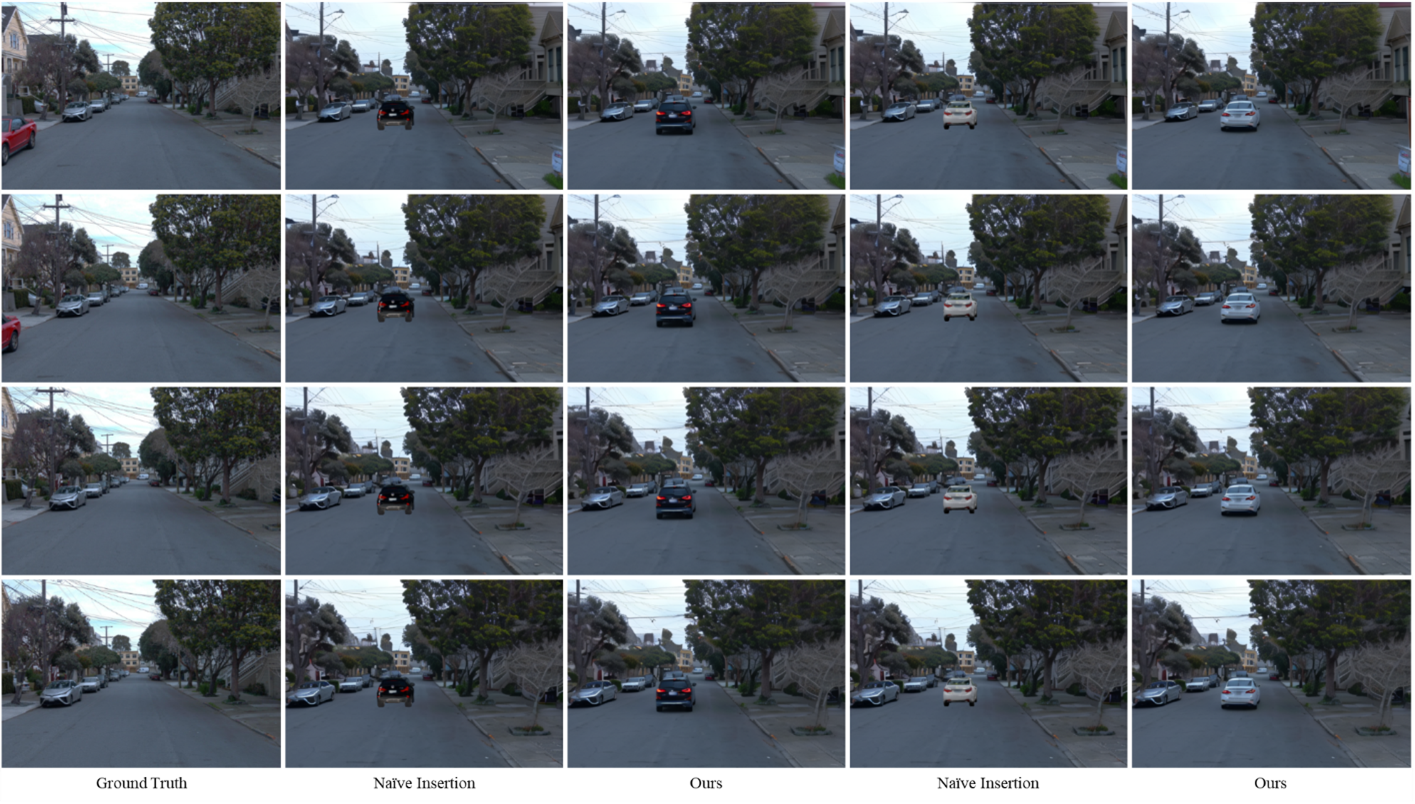}
\caption{Qualitative results from SCPainter on simultaneous asset insertion and NVS. SCPainter jointly performs NVS and asset insertion, producing temporally consistent views in which inserted assets are realistically integrated into the scene. Viewpoint is shifted 2.0m to the right of the original camera trajectory.} \label{fig4}
\end{figure}

\raggedbottom
\section{Conclusion}
In this work, we presented \textbf{SCPainter}, a unified framework for jointly performing realistic 3D asset insertion and NVS in autonomous driving videos. By leveraging scene-level geometry from per-pixel depth point clouds and 3D asset-level geometry from 3D GS, SCPainter produces photorealistic, temporally consistent videos under novel camera trajectories while effectively integrating inserted assets. Experiments on the Waymo Open Dataset along with quantitative and qualitative results demonstrate that our method improves insertion realism while maintaining NVS quality, enabling the generation of diverse and challenging driving scenarios for data-driven autonomous driving research.

\end{document}